\documentclass{article}

\usepackage{microtype}
\usepackage{graphicx}
\usepackage{subcaption}
\usepackage{booktabs} 
\usepackage[table]{xcolor}
\usepackage{hyperref}

\usepackage[accepted]{icml2026}

\usepackage{amsmath}
\usepackage{amssymb}
\usepackage{mathtools}
\usepackage{amsthm}

\usepackage[capitalize,noabbrev]{cleveref}

\theoremstyle{plain}

\theoremstyle{definition}

\theoremstyle{remark}

\usepackage{graphicx}
\usepackage{hyperref}
\usepackage{url}
\usepackage{array}
\usepackage{multirow}
\usepackage{booktabs} 
\usepackage{wrapfig}
\usepackage{hyperref}
\usepackage{url}
\newcolumntype{P}[1]{>{\centering\arraybackslash}p{#1}} 
\newcolumntype{R}[1]{>{\raggedleft\arraybackslash}p{#1}} 
\newcolumntype{L}[1]{>{\raggedright\arraybackslash}p{#1}}

\newlength\savewidth

\definecolor{lower}{RGB}{58,71,172}     
\definecolor{higher}{RGB}{166,67,58}    
\definecolor{myyellow}{RGB}{244,164,96}
\definecolor{myblue}{RGB}{65,105,225}   
\definecolor{mygreen}{RGB}{47,139,87}   
\definecolor{Highlight}{HTML}{39b54a}  
\definecolor{lightblue}{RGB}{51,153,204}

\usepackage[textsize=tiny]{todonotes}

\icmltitlerunning{Submission and Formatting Instructions for ICML 2026}

\begin{document}

\twocolumn[
  \icmltitle{PaSTel: Anchoring Histology in Spatial Transcriptomics via Multi-Scale Hierarchical Bio-Prior Contrastive Pretraining}



  \icmlsetsymbol{equal}{*}

  \begin{icmlauthorlist}
    \icmlauthor{Azim  Dehghani Amirabad}{equal,yyy}
    \icmlauthor{Junchao Zhu }{equal,yyy}
    \icmlauthor{Pushpak Pati }{yyy}
    \icmlauthor{Walid Abdelmoula}{yyy}
    \icmlauthor{Tommaso Mansi}{yyy}
    \icmlauthor{Rui Liao}{yyy}
  \end{icmlauthorlist}

  \icmlaffiliation{yyy}{Johnson \& Johnson Innovative Medicine}
  \icmlcorrespondingauthor{Azim Dehghani Amirabad}{adehghan@its.jnj.com}
  \icmlcorrespondingauthor{Rui Liao}{rliao2@its.jnj.com}

  \icmlkeywords{Machine Learning, ICML}

  \vskip 0.3in
]



\printAffiliationsAndNotice{}  

\begin{abstract}
Spatial transcriptomics (ST) links tissue morphology with molecular programs, motivating multimodal pretraining methods that align histology images with gene expression. However, existing approaches suffer from two key limitations: spatially informative gene selection is often dominated by ubiquitous housekeeping genes, leading to weakly discriminative representations, and independent spot–patch alignment fails to capture spatial dependencies that are critical for tissue organization. To address these challenges, we introduce PaSTel, a hierarchical multimodal pretraining framework that integrates biological priors at three levels. At the spot level, TF-IDF reweighting is used to identify spatially informative genes; at the functional level, curated KEGG pathways serve as anchors for encoding global biological semantics; and at the regional level, spatial clustering aggregates neighboring spots to model meso-scale tissue structure. Across multiple downstream tasks, PaSTel consistently outperforms existing vision and vision–omics encoders, demonstrating that incorporating multiscale biological priors yields more informative and transferable representations for spatial transcriptomics.
\end{abstract}

\section{Introduction}

Spatial transcriptomics (ST) has emerged as a transformative technology, offering a molecular atlas aligned with histological structure. ST has deepened our understanding of disease mechanisms and tumor microenvironmental heterogeneity~\citep{arora2023spatial,smith2024challenges,jones2024optimizing,dent2013her2}. However, its high cost and technical complexity hinder scalability in clinical and large-cohort settings~\citep{zhu2025asign,pineiro2022research}. In contrast, pathology slides are routinely capture morphological patterns associated with underlying expression programs~\citep{ash2021joint}, motivating vision-omics models that infer molecular profiles directly from histology images~\citep{he2020integrating,xie2023spatially}.

\begin{figure*}[t]
\centering
\includegraphics[width=0.8\linewidth]{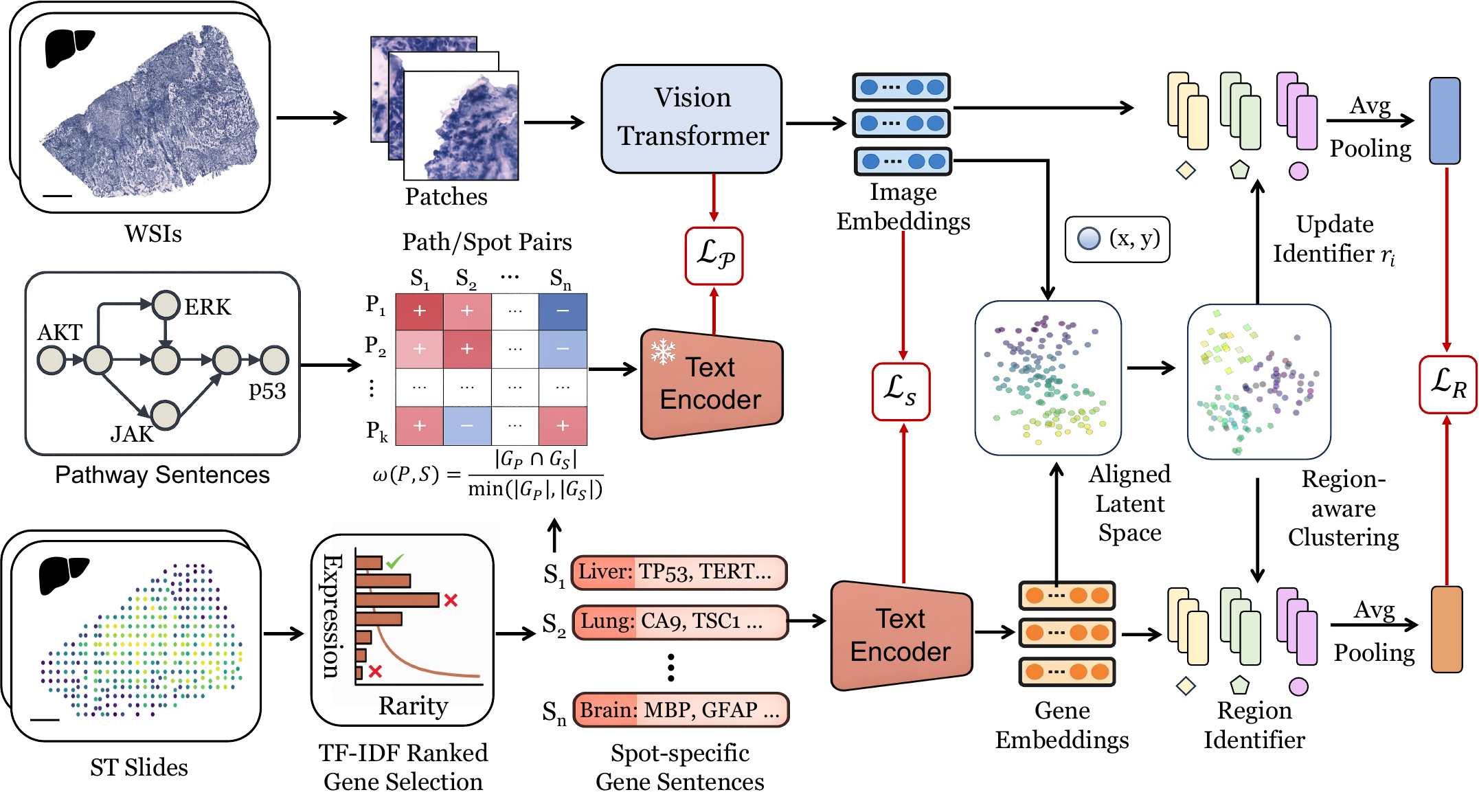}
  \caption{\textbf{Overview of PaSTel pretraining framework.}
  PaSTel adopts a progressive multi-level alignment strategy to harmonize visual features with molecular profiles.  First, at the local scale, vision-omics pairs are constructed by selecting informative ``keyword'' genes through a TF-IDF reweighting scheme to support fine-grained alignment. Then, curated pathway anchors are matched to each spot by computing gene-set overlaps, thereby defining positive and negative pairs at the functional scale, which serve as soft labels to inject biological knowledge. After initial pretraining at the spot-patch level, spatially aware clustering assigns region identifiers to contiguous spots, thereby integrating local patches into region-aware representations. Thus, PaSTel progressively incorporates local molecular signals and vision features with pathway-level semantics and meso-scale structures, achieving a unified multimodal representation.}
  \label{fig:framework}
\end{figure*}
Early vision-driven approaches are typically task-specific and trained on small, homogeneous datasets, limiting generalization across tissues, disease types, and sequencing platforms~\citep{he2020integrating,zeng2022spatial,pang2021leveraging,zhu2025diffusion,shi2024high}. To overcome this, recent multimodal pretraining frameworks adopt contrastive learning to align histology patches with gene expression, yielding more transferable vision-omics representations~\citep{chen2025visual}. Despite this progress, two key limitations remain. First, gene selection defines the gene “sentences” used for alignment and representation learning, yet common strategies are suboptimal. Top‑expressed genes are dominated by housekeeping signals \cite{stark2019rna,he2020integrating}, HVGs capture global rather than spatial variance \cite{stuart2019comprehensive}, and SVGs emphasize smooth patterns while missing sparse local markers \cite{sun2023spatial}, resulting in weak local discriminability and limited spot‑level representations.
Second, existing methods align spot–patch pairs independently, overlooking spatial dependencies that govern interactions and regional specialization, thus failing to capture multiscale biological structure.

To address these issues, we propose \textbf{PaSTel}, a hierarchical multimodal pretraining framework that injects biological priors into vision–language alignment for spatial transcriptomics. PaSTel introduces three complementary levels of supervision reflecting biological hierarchy. At the \emph{spot level}, a TF-IDF reweighting scheme suppresses ubiquitous genes while emphasizing spatially informative ``keywords,'' producing more discriminative representations. At the \emph{functional level}, curated KEGG pathways act as biological anchors, whose prototype embeddings are softly aligned with histology features via overlap-weighted contrastive learning to encode global functional semantics. At the \emph{regional level}, a spatially aware clustering module aggregates neighboring spots into coherent meso-scale regions during training, enabling modeling of long-range spatial dependencies.

By integrating these priors, PaSTel captures local molecular identity, global functional programs, and spatial tissue organization in a unified representation. The resulting vision-omics encoder is both biologically grounded and highly transferable, consistently outperforming existing pretrained vision and vision-omics models across benchmarks including gene expression prediction, few-shot learning, zero-shot spatial clustering, and image-to-ST sentence retrieval..

\noindent\textbf{Contributions.}
\begin{itemize}
    \item We propose a biologically informed alignment framework that combines TF-IDF-based gene selection with pathway-level supervision, improving the specificity and interpretability of spot representations.
    \item We introduce a region-aware hierarchical pretraining strategy that captures meso-scale spatial structure and enables multiscale modeling of tissue organization.
    \item We present PaSTel, a pretrained vision-omics encoder that achieves strong and consistent performance across diverse spatial transcriptomics tasks and datasets.
\end{itemize}

\begin{table*}[thbp] 
\centering
\scriptsize
\renewcommand{\arraystretch}{1} 
\begin{tabular}{
    p{0.077\linewidth}  
    p{0.14\linewidth}  
     P{0.052\linewidth} P{0.052\linewidth} P{0.052\linewidth}  
     P{0.052\linewidth} P{0.052\linewidth} P{0.052\linewidth}  
     P{0.052\linewidth} P{0.052\linewidth} P{0.052\linewidth}  
}

\toprule
\multirow{2}{*}{\textbf{Manner}} & \multirow{2}{*}{\textbf{Model}} & \multicolumn{3}{c}{\textbf{HER2}} & \multicolumn{3}{c}{\textbf{Breast Cancer}} & \multicolumn{3}{c}{\textbf{Kidney}} \\
\cline{3-11}
& & MSE & MAE & PCC & MSE & MAE & PCC & MSE & MAE & PCC \\
\midrule

\multirow{5}{*}{Regression}
& ST-Net   & 0.9225 & 0.7578 & 0.2829 & 0.6757 & 0.6656 & 0.1535 & 0.7636 & 0.6894 & 0.1551 \\
& HisToGene  & 0.9596 & 0.7944 & 0.2190 & 0.6905 & 0.6685 & 0.1113 & 0.8102 & 0.7158 & 0.1114 \\
& His2ST     & 1.0099 & 0.8194 & 0.0102 & 0.7084 & 0.6792 & 0.0019 & 0.8637 & 0.7437 & 0.0049 \\
& EGN        & 0.9473 & 0.7883 & 0.2266 & 0.6777 & 0.6362 & 0.1270 & 0.7811 & 0.6997 & 0.1512 \\
& TRIPLEX    & 0.9864 & 0.8031 & 0.0935 & 0.6834 & 0.6566 & 0.0421 & 0.7290 & 0.6757 & 0.0835 \\
\midrule

\multirow{5}{*}{Retrieval}
& BLEEP      & 0.8920 & 0.7602 & 0.2416 & 0.7729 & 0.7124 & 0.1149 & 0.8845 & 0.7669 & 0.1432 \\
& mclSTExp   & 0.9661 & 0.8705 & 0.1942 & 0.9451 & 0.7781 & 0.1013 & 1.0569 & 0.8366 & 0.1669 \\
& OpenCLIP   & 1.0560 & 0.7862 & 0.2780 & 0.6886 & 0.6517 & 0.1333 & 0.8737 & 0.7330 & 0.1833 \\
& OmiCLIP    & 0.8655 & 0.7986 & 0.2940 & 0.6260 & 0.6430 & 0.1930 & 0.7496 & 0.7097 & 0.2275 \\
 &  \cellcolor{orange!8} \underline{\textbf{PaSTel (Ours)}} & \cellcolor{orange!8} \textcolor{orange}{0.8298} & \cellcolor{orange!8} \textcolor{orange}{0.7298} & \cellcolor{orange!8} \textcolor{orange}{0.3172} & \cellcolor{orange!8} \textcolor{orange}{0.6082} & \cellcolor{orange!8} \textcolor{orange}{0.6074} & \cellcolor{orange!8} \textcolor{orange}{0.2480} & \cellcolor{orange!8} \textcolor{orange}{0.6971} & \cellcolor{orange!8} \textcolor{orange}{0.6749} & \cellcolor{orange!8} \textcolor{orange}{0.2627} \\
\bottomrule
\end{tabular}
\caption{\textbf{Quantitative comparisons on the gene expression prediction task.}  The best performance is highlighted in \textcolor{orange}{orange}, where we can observe that PaSTel outperforms the SOTAs across datasets.}
\label{tab:gene_prediction}
\end{table*}

\begin{figure*}[tbp]
\centering
\includegraphics[width=0.9\linewidth]{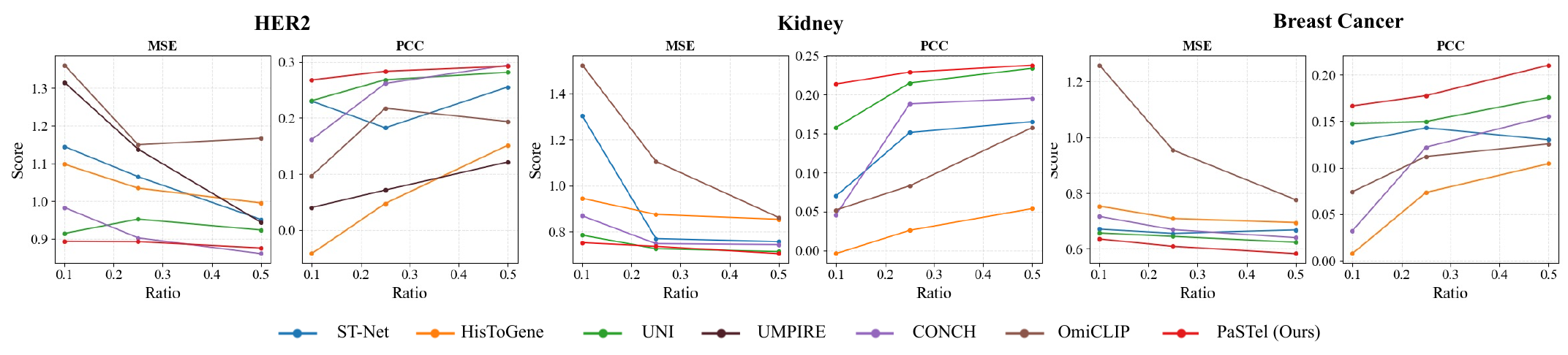}
\caption{\textbf{Few-shot gene expression prediction.} 
We evaluate performance under training data ratios of 10\%–50\% to simulate low-resource scenarios, with each curve denoting a model.
}
    \label{fig:few_shot}
\end{figure*}
\section{Methods}

We present PaSTel, a hierarchical multimodal pretraining framework that aligns histology images with gene expression by incorporating biological priors at three levels (Fig.~\ref{fig:framework}). PaSTel introduces supervision at spot, functional, and regional levels to jointly capture local molecular identity, global functional programs, and spatial tissue organization.

\subsection{TF-IDF Keyword Gene Selection}

A central challenge in vision--language alignment for spatial transcriptomics lies in constructing informative gene ``sentences''. Naively selecting top-expressed genes is dominated by ubiquitous housekeeping signals, leading to limited discrimination across spatial spots. To address this, we adopt a TF-IDF reweighting scheme that highlights genes that are both locally abundant and globally informative. Given an expression matrix $X \in \mathbb{R}^{B \times N_h}$, we define:
\begin{equation}
\begin{aligned}
\mathrm{TF}(g_i, s_j)
&= \frac{\log(1 + X_{j,i})}
{\sum_k \log(1 + X_{j,k}) + \epsilon}, \\
\mathrm{IDF}(g_i)
&= \log \frac{1 + M}
{1 + \sum_j \mathbf{1}[X_{j,i} > 0]} .
\end{aligned}
\end{equation}
The TF-IDF weight is then computed as:
\begin{equation}
w_{j,i}
= \mathrm{TF}(g_i, s_j)\cdot \mathrm{IDF}(g_i).
\end{equation}
We select the top-$K$ genes per spot to form a gene sentence $\mathcal{S}(s_j)$, yielding sparse and discriminative representations for multimodal alignment.

\subsection{Pathway-Level Soft Anchoring}

While TF-IDF captures local expression patterns, it does not explicitly model higher-level functional structure. To address this, we incorporate curated KEGG pathways as biological anchors. Given a pathway gene set $G_P$ and a spot gene set $G_S$, we define a normalized overlap score:
\begin{equation}
\omega(P,S) = \mathbf{1}\!\left[\frac{|G_P \cap G_S|}{\min(|G_P|,|G_S|)} > \rho\right]
\cdot \frac{|G_P \cap G_S|}{\min(|G_P|,|G_S|)}.
\end{equation}

Each pathway is represented by a prototype embedding:
\begin{equation}
\mathbf{z}_P = \frac{1}{|G_P|}\sum_{g \in G_P} \mathbf{e}_g.
\end{equation}

We use $\omega(P,S)$ as soft supervision to align pathway prototypes with vision embeddings, enabling model to encode global functional semantics with local expression patterns.

\subsection{Region-Level Aggregation}

To capture spatial dependencies beyond individual spots, we group neighboring spots into coherent regions. We define a hybrid distance that combines spatial proximity and feature similarity:
\begin{equation}
D_{ij} = \alpha \cdot d_{\text{spatial}}(\mathbf{x}_i,\mathbf{x}_j) + \beta \cdot d_{\text{feat}}([\mathbf{v}_i;\mathbf{t}_i],[\mathbf{v}_j;\mathbf{t}_j]).
\end{equation}

A $k$-nearest neighbor graph is constructed with weights $W_{ij} = \exp(-D_{ij})$, and Leiden clustering is applied to obtain region assignments $r_i$. The clustering is updated periodically during training, allowing progressively refined region-level supervision.

\subsection{Multi-Level Contrastive Objective}

We jointly optimize alignment across the three levels:
\begin{equation}
\mathcal{L} = \lambda_1 \mathcal{L}_s + \lambda_2 \mathcal{L}_p + \lambda_3 \mathcal{L}_r.
\end{equation}

At the spot level, we use a symmetric InfoNCE loss to align image and gene embeddings. At the pathway level, we match predicted similarities to soft targets derived from $\omega(P,S)$ using a KL-based objective. At the region level, we align region prototypes obtained by averaging embeddings within each cluster. We set $(\lambda_1,\lambda_2,\lambda_3) = (1, 0.25, 0.25)$.

\begin{figure*}[thpb]  
    \centering
    \includegraphics[width=0.9\linewidth]{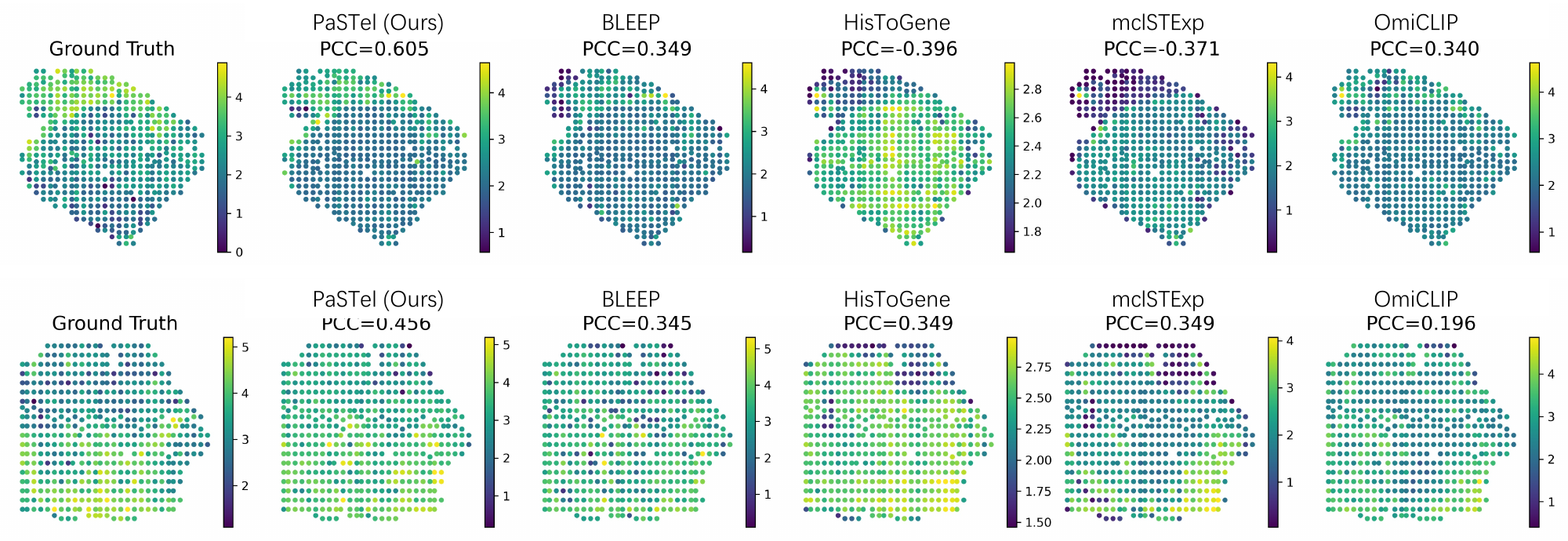}
    \caption{
\textbf{Visualization of the spatial expression distribution of the cancer-associated gene IGLL5.} The ground truth and predictions are shown for representative WSIs. Our PaSTel model achieves the highest PCC, accurately capturing spatial heterogeneity.
}
    \label{fig:biomarker}
\end{figure*}

\section{Data and Experiments}

\noindent \textbf{Dataset.}
We use HEST‑1K~\citep{jaume2024hest}, a large multi‑cohort spatial transcriptomics benchmark spanning diverse tissues and platforms. To prevent data leakage, we split HEST‑1K into disjoint pretraining and evaluation sets, holding out four cohorts—Breast Cancer~\citep{he2020integrating}, HER2~\citep{andersson2021spatial}, human DLPFC~\citep{maynard2021transcriptome}, and Kidney~\citep{lake2023atlas}—exclusively for downstream tasks. These datasets support zero‑shot clustering, gene expression prediction, few‑shot learning, and image‑to‑ST sentence retrieval, while all remaining Visium‑based cohorts are used only for pretraining.

\noindent \textbf{Implementation.}
We extract $224 \times 224$ image patches and select the top-250 TF-IDF genes per spot. The model adopts a ViT-B/16 image encoder and a transformer-based text encoder within OpenCLIP~\citep{ilharco_gabriel_2021_5143773}, treating each gene as an individual token. Training is performed with AdamW for 50 epochs on 8 A100 GPUs, with region-level aggregation enabled after epoch 25.

\noindent \textbf{Tasks.}
We evaluate four downstream tasks: few-shot gene prediction, zero-shot spatial clustering, gene reconstruction, and image-to-ST retrieval, covering predictive performance and representation transferability. Details are in appendix.

\section{Results}

\subsection{Low-Resource and Zero-Shot Evaluation}

We evaluate how pretrained encoders adapt to held-out ST datasets under limited supervision. Baselines include DenseNet121~\citep{huang2017densely} and pathology-pretrained encoders CONCH~\citep{lu2024avisionlanguage}, UNI~\citep{chen2024uni}, OmiCLIP~\citep{chen2025visual}, and UMPIRE~\citep{han2025towards}, all trained under a unified protocol for fair comparison. As shown in Fig.~\ref{fig:few_shot}, PaSTel consistently outperforms baselines across varying training data ratios, with largest gains observed in low-data regimes (e.g., 10\%), where learning reliable vision-omics mappings is most challenging. This suggests that hierarchical biological priors lead to generalizable representations under limited supervision.

We further assess representation quality using vision-only zero-shot clustering on DLPFC and HER2 datasets. Without finetuning or access to gene expression, PaSTel achieves the best performance on both laminar structures and tumor regions, indicating that it learns spatially coherent and semantically meaningful visual representations. Additional results are shown in Table~\ref{tab:nosmooth_zero_shot} and Figure~\ref{fig:sup_cluster_her2}.

\subsection{Validation on Gene Expression Prediction}

\noindent \textbf{Cross-Validation.} We compare PaSTel with both regression-based (ST-Net~\citep{he2020integrating}, EGN~\citep{yang2023exemplar}, HisToGene~\citep{pang2021leveraging}, His2ST~\citep{zeng2022spatial}, TRIPLEX~\citep{chung2024accurate}) and retrieval-based methods (BLEEP~\citep{xie2024spatially}, mclSTExp~\citep{min2024multimodal}, OpenCLIP~\citep{ilharco_gabriel_2021_5143773}, OmiCLIP~\citep{chen2025visual}). We predict the top 300 highly variable genes (HVGs) and report PCC, MSE, and MAE. 

As shown in Table~\ref{tab:gene_prediction}, PaSTel consistently achieves the best performance across all datasets. Improvements are particularly pronounced on the Kidney cohort, where PaSTel attains a PCC of 0.2627, outperforming OmiCLIP (0.2275) and substantially exceeding regression-based baselines. This highlights its superior generalization in heterogeneous tissues, where biological priors provide complementary signals beyond morphology alone.

\noindent \textbf{Biomarker Prediction.}
We further evaluate spatial prediction of the clinically relevant gene IGLL5. As shown in Fig.~\ref{fig:biomarker}, PaSTel achieves the highest PCC (0.605 and 0.456) and accurately recovers fine-grained spatial expression patterns, while competing methods yield weak or even negative correlations. These results demonstrate that PaSTel captures meaningful spatial variation beyond coarse trends.

\section{Conclusion}
We presented \textbf{PaSTel}, a hierarchical multimodal pretraining framework that integrates biological priors into vision–language alignment for spatial transcriptomics through TF-IDF-based gene selection, pathway-level anchoring, and region-level aggregation. Across a range of downstream tasks, PaSTel consistently outperforms strong vision and vision-omics baselines, with the most pronounced improvements under data-scarce and heterogeneous tissue settings. Ablation studies further demonstrate that each component contributes meaningfully, underscoring the importance of modeling multiscale biological structure for learning robust and transferable representations.

\nocite{langley00}

\bibliography{example_paper}
\bibliographystyle{icml2026}

\newpage
\appendix
\onecolumn

\section{Appendix: Detailed Method}

\subsection{TF-IDF Formulation Details}

We apply a log-transformation $\log(1+x)$ to stabilize the variance of count-based gene expression data. The term frequency (TF) measures the relative prominence of a gene within a spot, while the inverse document frequency (IDF) captures how rare a gene is across all spots:
\begin{equation}
\mathrm{TF}(g_i, s_j) = \frac{\log(1 + X_{j,i})}{\sum_{k=1}^{N_h} \log(1 + X_{j,k}) + \epsilon},
\end{equation}
\begin{equation}
\mathrm{IDF}(g_i) = \log \left( \frac{1 + M}{1 + \sum_{j=1}^{M} \mathbf{1}[X_{j,i} > 0]} \right),
\end{equation}
where $\mathbf{1}[\cdot]$ is the indicator function. Genes that are frequently expressed across many spots receive lower IDF weights, while spatially specific genes are emphasized.

The TF-IDF weight is defined as:
\begin{equation}
w_{j,i} = \mathrm{TF}(g_i, s_j) \cdot \mathrm{IDF}(g_i),
\end{equation}
and gene sentences are constructed by selecting the top-$K$ genes per spot based on $w_{j,i}$.

\subsection{Pathway Supervision Details}

We use curated KEGG pathways as biological priors to provide functional supervision. Each pathway $P$ is associated with a gene set $G_P \subseteq \mathcal{G}$. Given a spot-specific gene set $G_S$, we compute the normalized overlap:
\begin{equation}
\omega(P, S) = \mathbf{1}\!\left[ \frac{|G_P \cap G_S|}{\min(|G_P|, |G_S|)} > \rho \right] 
\cdot \frac{|G_P \cap G_S|}{\min(|G_P|, |G_S|)}.
\end{equation}

The threshold $\rho$ filters low-relevance pathway-spot pairs, preventing noisy supervision. Because each spot may participate in multiple biological processes, we treat $\omega(P,S)$ as a soft supervision signal rather than a binary label.

Each pathway is represented by a prototype embedding obtained via mean pooling:
\begin{equation}
\mathbf{z}_P = \frac{1}{|G_P|} \sum_{g \in G_P} \mathbf{e}_g,
\end{equation}
which provides a robust summary of pathway-level functional semantics.

\subsection{Region Clustering Details}

To model spatial structure, we define a distance that combines spatial proximity and multimodal similarity. The spatial distance is computed using Euclidean coordinates, while the feature distance is based on cosine dissimilarity of concatenated image and gene embeddings:
\begin{equation}
D_{ij} = \alpha \cdot 
\frac{\|\mathbf{x}_i - \mathbf{x}_j\|_2 - d_{\min}}{d_{\max} - d_{\min}}
+ \beta \cdot 
\frac{1 - \cos([\mathbf{v}_i;\mathbf{t}_i], [\mathbf{v}_j;\mathbf{t}_j]) - c_{\min}}{c_{\max} - c_{\min}}.
\end{equation}

Here, $\mathbf{x}_i$ denotes spatial coordinates, and $[\mathbf{v}_i;\mathbf{t}_i]$ is the concatenated multimodal embedding. The normalization terms $d_{\min}, d_{\max}, c_{\min}, c_{\max}$ ensure both components lie in $[0,1]$. We set $\alpha = \beta = 0.5$.

A weighted $k$-nearest neighbor graph is constructed with weights:
\begin{equation}
W_{ij} = \exp(-D_{ij}),
\end{equation}
and Leiden clustering is applied to maximize modularity:
\begin{equation}
Q = \frac{1}{2m} \sum_{i,j} \left[ W_{ij} - \frac{k_i k_j}{2m} \right] \cdot \mathbf{1}[r_i = r_j],
\end{equation}
where $k_i = \sum_j W_{ij}$ and $m = \frac{1}{2}\sum_{i,j} W_{ij}$. This yields region assignments $r_i$ for each spot. Clustering is recomputed periodically during training to reflect updated embeddings.

\subsection{Multi-Level Loss Details}

The spot-level loss is defined using symmetric InfoNCE:
\begin{equation}
\mathcal{L}_{\mathbf{v} \rightarrow \mathbf{t}} = -\frac{1}{B} \sum_{i=1}^{B} 
\log \frac{\exp(\mathbf{v}_i^\top \mathbf{t}_i / \tau)}{\sum_{j=1}^{B} \exp(\mathbf{v}_i^\top \mathbf{t}_j / \tau)},
\end{equation}
\begin{equation}
\mathcal{L}_s = \tfrac{1}{2}(\mathcal{L}_{\mathbf{v} \rightarrow \mathbf{t}} + \mathcal{L}_{\mathbf{t} \rightarrow \mathbf{v}}).
\end{equation}

For pathway supervision, we construct a soft target distribution:
\begin{equation}
q_i(j) = \frac{\exp(\omega_{ij})}{\sum_{k} \exp(\omega_{ik})},
\end{equation}
and a predicted distribution:
\begin{equation}
\hat{q}_i(j) = \frac{\exp(\cos(\mathbf{v}_i, \mathbf{z}_{p_j}) / \tau_p)}{\sum_{k} \exp(\cos(\mathbf{v}_i, \mathbf{z}_{p_k}) / \tau_p)}.
\end{equation}

We minimize a saturating KL divergence:
\begin{equation}
\mathcal{L}_p = \frac{1}{B} \sum_{i=1}^{B} 
\left[ 1 - \exp\left( -\frac{1}{\gamma} \mathrm{KL}(\hat{q}_i \| q_i) \right) \right].
\end{equation}

At the region level, we define prototypes:
\begin{equation}
\bar{\mathbf{v}}_r = \frac{1}{|r|} \sum_{i \in r} \mathbf{v}_i, \quad
\bar{\mathbf{t}}_r = \frac{1}{|r|} \sum_{i \in r} \mathbf{t}_i,
\end{equation}
and apply a symmetric InfoNCE loss between region-level embeddings.

\subsection{Training Details}

Region assignments are updated periodically during training to reflect evolving representations. Temperature scaling is applied to both spot-level and pathway-level objectives to stabilize optimization.

\subsection{Image-to-ST Sentence Retrieval}

Image-to-sentence retrieval requires the model to identify the gene expression description that corresponds to a given histology image, providing a direct test of the semantic consistency between morphological features and molecular profiles. We evaluate retrieval quality with Recall@K\% on the Breast Cancer and Kidney datasets.

Table~\ref{tab:retrieval_results} reports retrieval performance of different models. PaSTel substantially outperforms existing methods on both cohorts. On Breast Cancer it lifts Recall@5\% from 15.63 (OmiCLIP, the strongest medical-pretrained baseline) to 19.88, and on Kidney it ranks first or second across every recall threshold. A key driver of this gap is input length. Medical foundation models such as OmiCLIP and PLIP are restricted to 77-token inputs, whereas PaSTel processes gene sentences of up to 250 tokens. The longer context lets the encoder reason over a substantially larger candidate gene pool while staying robust to the noise that extended sequences typically introduce.

Length alone is not enough. We additionally redesign the text vocabulary so that each gene name is mapped to a single atomic token rather than fragmented by subword schemes such as BPE. Under standard BPE tokenization, a 77-token budget typically encodes only 20--30 distinct genes, since each multi-character gene symbol consumes several subword pieces. The atomic-gene vocabulary preserves gene identity end-to-end and lets PaSTel index a much broader gene set, including lowly expressed but biologically informative markers that would otherwise be unreachable.

\subsection{Ablation Study}
We ablate PaSTel along three axes: the contribution of each functional block, the gene-sentence token length, and the two hyperparameters governing the biological priors. Unless otherwise stated, all settings other than the manipulated variable follow the default configuration.

\begin{table*}[thbp] 
\centering
\scriptsize
\renewcommand{\arraystretch}{1} 
\begin{tabular}{
    p{0.18\linewidth}  
     P{0.055\linewidth} P{0.055\linewidth} P{0.055\linewidth}  
     P{0.055\linewidth} P{0.055\linewidth} P{0.055\linewidth}  
     P{0.055\linewidth} P{0.055\linewidth} P{0.055\linewidth}  
}

\toprule
\multirow{2}{*}{\textbf{Functional Blocks}} & \multicolumn{3}{c}{\textbf{HER2}} & \multicolumn{3}{c}{\textbf{Breast Cancer}} & \multicolumn{3}{c}{\textbf{Kidney}} \\
\cline{2-10}
 & MSE & MAE & PCC & MSE & MAE & PCC & MSE & MAE & PCC \\
\midrule

w/o Pathway            & 0.8531 & 0.7626 & 0.2770 & 0.7031 & 0.7034 & 0.1689 & 0.7926 & 0.7112 & 0.2500 \\
w/o Regional Alignment & 0.8151 & 0.7407 & 0.3035 & 0.6501 & 0.6794 & 0.2162 & 0.7479 & 0.7006 & 0.2419 \\
Top Selection          & 0.9025 & 0.8011 & 0.2600 & 0.7833 & 0.7307 & 0.1481 & 0.8187 & 0.7385 & 0.2087 \\
HVG Selection          & 0.9563 & 0.8261 & 0.2580 & 0.7059 & 0.6817 & 0.1940 & 0.8344 & 0.7012 & 0.1877 \\
\rowcolor{orange!12}\underline{\textbf{PaSTel (Ours)}}        & \textbf{0.8298} & \textbf{0.7298} & \textbf{0.3172} & \textbf{0.6082} & \textbf{0.6074} & \textbf{0.2480} & \textbf{0.6971} & \textbf{0.6749} & \textbf{0.2627}  \\
\midrule
\midrule
Token Length=100       & 0.8250 & 0.7351 & 0.3060 & 0.6535 & 0.6968 & 0.2234 & 0.7290 & 0.6920 & 0.2620 \\
Token Length=500       & 0.8084 & 0.7374 & 0.3379 & 0.6406 & 0.6919 & 0.2056 & 0.6816 & 0.6694 & 0.2730 \\
\midrule
\midrule
Pathway Ratio $\rho$=0.1 & 0.8803 & 0.7647 & 0.2776 & 0.6494 & 0.6705 & 0.2396 & 0.7241 & 0.6951 & 0.2428 \\
Pathway Ratio $\rho$=0.5 & 0.8496 & 0.7532 & 0.3032 & 0.5475 & 0.5712 & 0.2589 & 0.7178 & 0.6637 & 0.2384 \\
\midrule
\midrule
Loss: $\lambda_2$=0.25, $\lambda_3$=0.1
& 0.8897 & 0.7673 & 0.2854
& 0.6359 & 0.6662 & 0.2615
& 0.7210 & 0.6653 & 0.2358 \\
Loss: $\lambda_2$=0.5, $\lambda_3$=0.5
& 0.8517 & 0.7521 & 0.2781
& 0.5895 & 0.6067 & 0.2636
& 0.7193 & 0.6636 & 0.2410 \\
\bottomrule
\end{tabular}
\caption{\textbf{Ablation study on gene expression prediction.} The default \textbf{PaSTel} configuration (highlighted row) is used as the reference; each subsequent block manipulates a single design choice while keeping all other settings fixed.}
\label{tab:ablation_tokenlength}
\end{table*}

\begin{table*}[htbp]
\centering
\scriptsize
\renewcommand{\arraystretch}{1} 
\begin{tabular}{p{0.13\linewidth} P{0.11\linewidth}   P{0.086\linewidth} P{0.086\linewidth} P{0.09\linewidth} P{0.086\linewidth} P{0.086\linewidth} P{0.09\linewidth}}
\toprule

\multirow{2}{*}{\textbf{Model}} & \multirow{2}{*}{\textbf{Token Length}}  & \multicolumn{3}{c}{\textbf{Breast Cancer}} & \multicolumn{3}{c}{\textbf{Kidney}} \\
\cline{3-8}
 & & Recall@1\% & Recall@5\% & Recall@10\% & Recall@1\% & Recall@5\% & Recall@10\% \\
\midrule

Random & -  & 0.792 & 5.032 & 8.797 & 0.928 & 4.950 & 9.171 \\
PLIP & 77 & 2.961 & 11.184 & 24.513 & 2.004 & 9.198 &  \textcolor{orange}{23.963}\\
OmiCLIP & 77 & \textcolor{lightblue}{\underline{3.397}} & \textcolor{lightblue}{\underline{15.630}} & \textcolor{lightblue}{\underline{28.093}} & \textcolor{orange}{2.347} & \textcolor{lightblue}{\underline{9.930}} & 17.799 \\
CONCH & 128 & 2.184 & 9.669 & 17.587 & 1.469 & 7.822 & 14.919 \\
OpenCLIP & 250 & 1.303 & 8.129 & 16.563 & 1.399 & 6.600 & 12.459 \\
\rowcolor{orange!8} \textbf{\underline{PaSTel (Ours)}}& 250 & \textcolor{orange}{3.597} & \textcolor{orange}{19.882} & \textcolor{orange}{32.432} & \textcolor{lightblue}{\underline{2.118}} & \textcolor{orange}{10.280} & \textcolor{lightblue}{\underline{19.352}} \\
\bottomrule

\end{tabular}
\caption{\textbf{Image-to-ST sentence retrieval performance across datasets}, with best performance highlighted in \textcolor{orange}{\textbf{orange}} and second highest in \textcolor{lightblue}{\underline{blue}}.}
\label{tab:retrieval_results}
\end{table*}

\noindent \textbf{Biological Information Prior.} We assess the contribution of each functional block by individually removing pathway-guided supervision and region-level alignment, and by replacing the TF-IDF gene sentence with two alternative gene-selection strategies: (i) Top Selection, which keeps the top genes ranked by raw expression at each spot, and (ii) HVG Selection, which uses a slide-level high-variance gene set shared by all spots. Token length is fixed at 250 across all settings. As shown in Tables~\ref{tab:ablation_tokenlength} and~\ref{tab:retrieval_ablation}, every component contributes to the final performance. Replacing TF-IDF with Top Selection or HVG Selection causes sizable drops across all three cohorts. For instance on HER2 where PCC falls from 0.3172 to 0.2600 (Top) and 0.2580 (HVG), confirming that spot-specific gene weighting is what makes the priors effective rather than the choice of any informative gene set. Removing pathway supervision likewise reduces prediction quality, and the drop is most pronounced on Breast Cancer (PCC 0.2480 $\to$ 0.1689) where pathway-level semantics carry more weight; the effect is milder but still consistent on HER2 and Kidney. The retrieval task amplifies these gaps: on Breast Cancer, Recall@5\% falls from 19.88 to 12.72 under Top Selection and to 17.16 without regional alignment, indicating that the biological priors are essential for robust cross-modal alignment.

\begin{table*}[htbp]
\centering
\scriptsize
\renewcommand{\arraystretch}{1} 
\begin{tabular}{p{0.18\linewidth}   P{0.08\linewidth} P{0.08\linewidth} P{0.09\linewidth} P{0.08\linewidth} P{0.08\linewidth} P{0.09\linewidth}}
\toprule
\multirow{2}{*}{\textbf{Functional Blocks}} & \multicolumn{3}{c}{\textbf{Breast Cancer}} & \multicolumn{3}{c}{\textbf{Kidney}} \\
\cline{2-7}
 & Recall@1\% & Recall@5\% & Recall@10\% & Recall@1\% & Recall@5\% & Recall@10\% \\
\midrule
w/o Pathway            & 2.180 & 12.591 & 26.785 & 1.820 & 8.911 & 17.523 \\
w/o Regional Alignment & 2.834 & 17.156 & 30.667 & 1.861 & 9.276 & 18.130 \\
Top Selection          & 2.407 & 12.725 & 24.703 & 1.672 & 7.580 & 15.096 \\
HVG Selection          & 1.920 & 8.675  & 25.674 & 1.395 & 7.796 & 15.254 \\
\rowcolor{orange!12}\textbf{\underline{PaSTel (Ours)}}      & \textbf{3.597} & \textbf{19.882} & \textbf{32.432} & \textbf{2.118} & \textbf{10.280} & \textbf{19.352} \\
\bottomrule
\end{tabular}
\caption{\textbf{Ablation study on image-to-sentence retrieval.} The default \textbf{PaSTel} configuration (highlighted row) is the reference; each other row manipulates a single design choice while keeping all remaining settings fixed.}
\label{tab:retrieval_ablation}
\end{table*}

\begin{figure}[htbp]
    \centering
    \includegraphics[width=0.9\linewidth]{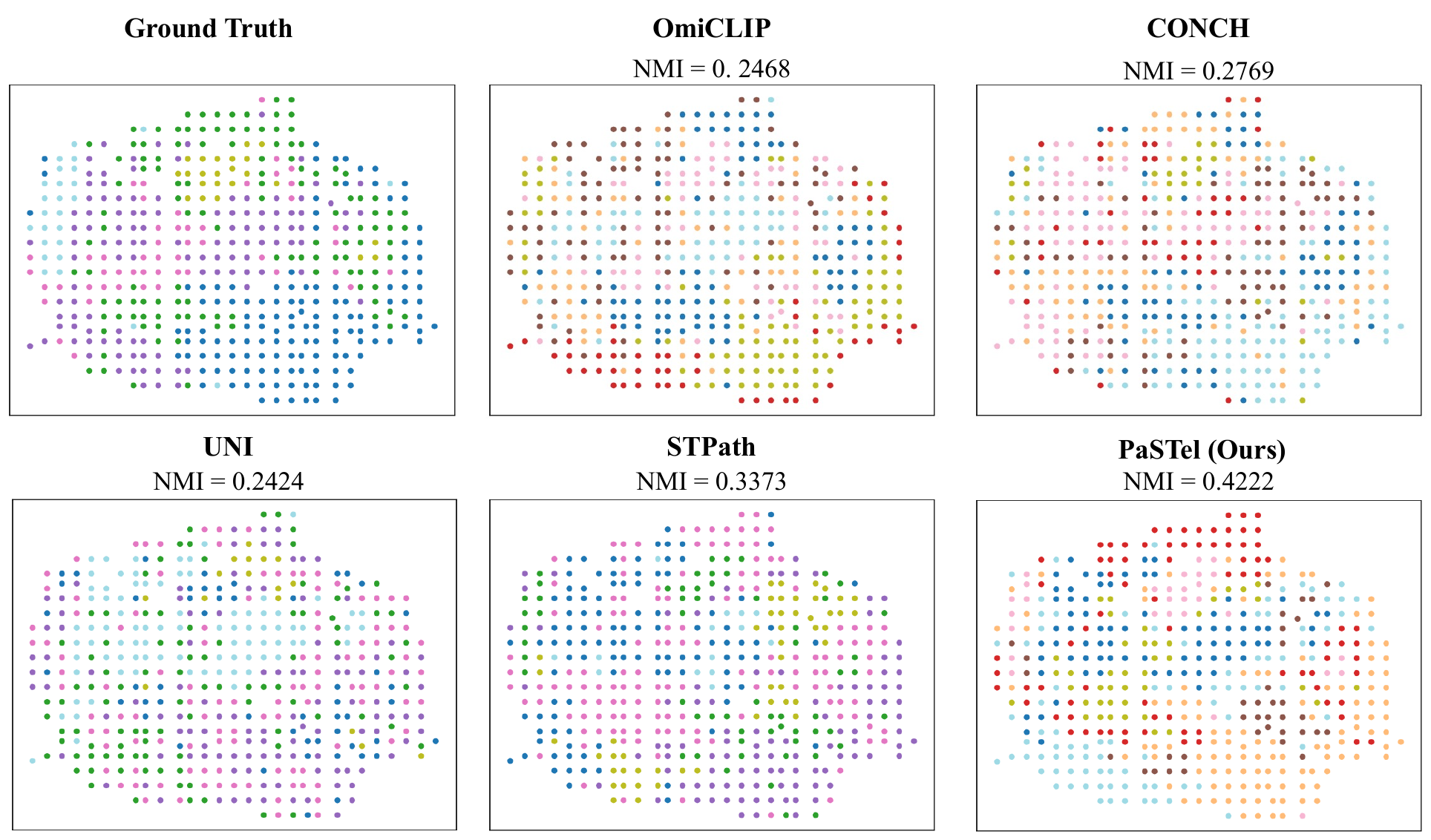}
    \caption{Visualization of zero-shot spatial clustering results on the HER2 dataset.}
    \label{fig:sup_cluster_her2}
\end{figure}

\noindent \textbf{Hierarchical Regional Alignment.} Removing the region-aggregation branch keeps spot-level alignment intact but discards the meso-scale signal. Tables~\ref{tab:ablation_tokenlength} and~\ref{tab:retrieval_ablation} show that this hurts both prediction (PCC drops from 0.2480 to 0.2162 on Breast Cancer) and retrieval (BC Recall@5\% from 19.88 to 17.16). The supplementary clustering experiments (Table~\ref{tab:ablation_clustering} further show that regional alignment is the single component most responsible for the model's ability to recover spatially coherent tissue layers without supervision, suggesting that local contrastive alignment alone is insufficient to instill a region-aware geometry into the visual feature space. 

\begin{table}[t]
\centering
\scriptsize
\begin{tabular}{lcc}
\toprule
\textbf{Model} & \textbf{ARI} & \textbf{NMI} \\
\midrule
OmiCLIP              & 0.1220 & 0.1747 \\
CONCH                & 0.1346 & 0.2070 \\
UNI                  & 0.1322 & 0.2127 \\
ST-Path              & \textcolor{lightblue}{\underline{0.1695}} & \textcolor{lightblue}{\underline{0.2568}} \\
\textbf{\underline{PaSTel (Ours)}} & \textcolor{orange}{0.2503} & \textcolor{orange}{0.3542} \\
\bottomrule
\end{tabular}
\caption{Zero-shot clustering performance without smoothing on the DLPFC dataset across foundation models.}
\label{tab:nosmooth_zero_shot}
\end{table}


\begin{table}[htbp]
\centering
\scriptsize
\renewcommand{\arraystretch}{1} 
\begin{tabular}{
    p{0.15\linewidth}     
    p{0.20\linewidth}      
    P{0.1\linewidth}      
    P{0.1\linewidth}      
}
\toprule
\textbf{Category} & \textbf{Setting} & \textbf{ARI} & \textbf{NMI} \\
\midrule
\multirow{3}{*}{Functional Block}
& Top Selection           & 0.1861 & 0.3035 \\
& w/o Regional Align      & 0.2262 & 0.3252 \\
& PaSTel (Ours)           & 0.2640 & 0.3776 \\
\midrule
\midrule
\multirow{3}{*}{Pathway Ratio}
& $\rho = 0.1$            & 0.2288 & 0.3458 \\
& $\rho = 0.2$            & 0.2640 & 0.3776 \\
& $\rho = 0.5$            & 0.2399 & 0.3703 \\
\midrule
\midrule
\multirow{4}{*}{Loss Weights}
& $\lambda_2 = 0, \lambda_3=0$              & 0.2092 & 0.3112 \\
& $\lambda_2 = 0.25, \lambda_3=0.1$         & 0.2306 & 0.3512 \\
& $\lambda_2 = 0.25, \lambda_3=0.25$        & 0.2640 & 0.3776 \\
& $\lambda_2 = 0.5, \lambda_3=0.5$          & 0.2678 & 0.3871 \\
\midrule
\midrule
HVG Selection
& Token Length=250              & 0.1573 & 0.2726 \\
\bottomrule
\end{tabular}
\caption{Ablation study of zero-shot clustering performance on DLPFC dataset.}
\label{tab:ablation_clustering}
\end{table}

\noindent \textbf{Pathway Overlap Ratio.} The threshold $\rho$ controls how strictly two pathways must share their gene composition before being treated as a positive pair in the contrastive loss. We compare $\rho = 0.1$ and $\rho = 0.5$ against the default $\rho = 0.2$ in Tables~\ref{tab:ablation_tokenlength}. A loose threshold ($\rho = 0.1$) floods the positive set with biologically unrelated pathways and degrades performance across all three cohorts, while a strict threshold ($\rho = 0.5$) depletes anchor coverage and hurts the more heterogeneous HER2 and Kidney cohorts where the loss becomes insufficiently supervised. The default $\rho = 0.2$ delivers the most consistent performance across cohorts and is therefore retained as the standard configuration.

\noindent \textbf{Loss Weights Analysis.} The weights $\lambda_2$ and $\lambda_3$ control the relative emphasis placed on pathway supervision and regional alignment over the spot-level loss. As shown in Tables~\ref{tab:ablation_tokenlength}, lowering $\lambda_3$ from the default 0.25 to 0.1 (with $\lambda_2$ kept at 0.25) noticeably degrades HER2 and Kidney performance, with PCC dropping from 0.3172 to 0.2854 on HER2 and from 0.2627 to 0.2358 on Kidney, confirming that the regional signal is load-bearing. Raising both weights to 0.5 also fails to consistently surpass the default across cohorts and additionally introduces less stable training dynamics. The default $(\lambda_2, \lambda_3) = (0.25, 0.25)$ thus offers the most balanced behavior across all three cohorts.

\noindent \textbf{Token Length of Gene Sentence.} We sweep the gene-sentence length over $\{100, 250, 500\}$ tokens, with results in Table~\ref{tab:ablation_tokenlength}. A short sentence (100 tokens) systematically loses on Breast Cancer and Kidney because the truncation drops too many discriminative genes per spot. Extending to 500 tokens admits low-expression genes that act as noise on Breast Cancer (MSE 0.6082 $\to$ 0.6406) and provides no consistent gain across the three cohorts. The default of 250 tokens delivers the most balanced behavior and is therefore retained as our standard configuration.

\end{document}